\documentclass[12pt]{article}
\usepackage{graphicx} % Required for inserting images
\usepackage{amsmath}
\usepackage[
    style = authoryear,
    isbn=false,
    dashed=false,
    mincitenames = 1,
    maxcitenames = 2,
    maxbibnames=5,
    minbibnames=5,
    uniquelist=false,
    uniquename=false
]{biblatex}
\usepackage[T1]{fontenc}
\usepackage{fullpage}
\usepackage{todonotes}
\usepackage{multirow}
\usepackage{amsfonts}
\usepackage{setspace}
\usepackage{tipa}
\usepackage{xurl}
\usepackage{hyperref}
\usepackage{authblk}
\DeclareFieldFormat[article,inbook,incollection,inproceedings,patent,thesis,unpublished]{title}{#1}
\DeclareDelimFormat{nameyeardelim}{\addcomma\space}
\DeclareFieldFormat{postnote}{#1}
\DeclareFieldFormat{multipostnote}{#1}

\AtEveryBibitem{%
  \clearfield{note}%
  \clearlist{language}%
}

\title{Investigating Lexical Change through Cross-Linguistic Colexification Patterns}

\author{Kim Gfeller}
\author{Sabine Stoll}
\author{Chundra Cathcart}
\author{Paul Widmer}

\affil{Institute for the Interdisciplinary Study of Language Evolution\\
University of Zurich}

\date{}

\begin{document}
%\tableofcontents

\onehalfspacing

\maketitle

\begin{abstract}
    \noindent One of the most intriguing features of language is its constant change, with ongoing shifts in how meaning is expressed. Despite decades of research, the factors that determine how and why meanings evolve remain only partly understood. 
    %Colexification --- the phenomenon of expressing multiple distinct concepts using the same word form --- provides a valuable concept for investigating how lexical meanings change across languages. 
    Colexification --- the phenomenon of expressing multiple distinct concepts using the same word form --- serves as a valuable window onto the dynamics of meaning change across languages. 
    %Here, we apply phylogenetic comparative models to dictionary data from three language families, Austronesian, Indo-European, and Uralic, to estimate change parameters of the colexification of concept pairs. 
    Here, we apply phylogenetic comparative models to dictionary data from three language families, Austronesian, Indo-European, and Uralic, in order to shed light on the evolutionary dynamics underlying the colexification of concept pairs. 
    We assess the effects of three predictors: associativity, borrowability, and usage frequency. Our results show that more closely related concept pairs are colexified across a larger portion of the family tree and exhibit slower rates of change. 
    In contrast, concept pairs that are more frequent and more prone to borrowing tend to change more rapidly and are less often colexified. We also find considerable differences between the language families under study, suggesting that areal and cultural factors may play a role.
\end{abstract}

%\textbf{Keywords}: colexification, lexical change, phylogenetic comparative methods, meaning change, Bayesian modeling

\section{Introduction}
All languages name and categorize aspects of human experience, but they do so in different ways. For instance, some languages use the same word for the concepts \textsc{hand} and \textsc{arm}, while others use distinct words. Such differences arise through processes of lexical change, in which words gain, lose, or shift meanings over time. Understanding how these processes unfold is important because they shape the categories speakers use to communicate and reflect broader cultural and historical dynamics as well as cognitive predispositions. Yet, the factors which determine whether distinctions between concepts are maintained or lost in the lexicon remain unclear.

Quantitative studies of lexical change face three key challenges. First, the range of possible lexical changes is vast \parencite{urban_lexical_2014}. Various attempts have been made to categorize these changes. \textcite{DatSemShift}, for example, provide a database of attested semantic shifts, while \textcite{traugott_regularity_2001} focus on some widely attested examples of meaning change. A more semantically restricted approach to regularity in change is the field of grammaticalization \parencite{heine_world_2002}, which examines regular semantic shifts from less grammatical to more grammatical meanings, such as English demonstrative \textit{that} to more grammatical subordinating \textit{that}. However, these frameworks deal only with attested shifts; therefore, they may not capture the full variety of changes, as many shifts might have occurred before languages were adequately documented. These approaches are also largely static, as they rely on discrete, observable changes and do not model the underlying processes over time. Moreover, they do not account for the fact that languages are not statistically independent due to phylogenetic autocorrelation, a challenge known as Galton's problem \parencite{bromhamSolvingGaltonsProblem2024}.

The second challenge for quantitative studies of lexical change is the difficulty of tracking meaning change at the level of individual lexemes. This is particularly problematic for phylogenetic studies, which often deal with large time depths \parencite{bowern-2019-semantic}. The difficulty arises because lexemes are not stable over time: their morphological form may change, be replaced, or be borrowed from another language. Consequently, assuming a fixed lexical item across extended time depths is often unrealistic.

Finally, research on meaning change is often limited in the number of languages or in the time span under investigation. For example, \textcite{xu2017evolution} study metaphorical extensions of concepts over a millennium, but only in English, because comparable data are not available for most other languages. Conversely, \textcite{hamilton2016diachronic} use word embedding data from four languages from the last 200 years, because such data are difficult to obtain for longer time spans and more stages of languages.

To address these challenges, we turn to colexification, the expression of distinct concepts by a single word form \parencite{francois_semantic_2008}. Even closely related languages differ in their colexification patterns: while English uses two distinct forms for the concepts \textsc{husband} and \textsc{man}, German uses \textit{Mann} for both. Such differences reflect lexical change and provide a systematic window into how meanings shift. Colexification can be conceptualized as the presence or absence of two concepts being expressed by the same linguistic form, regardless of what that form is. This is crucial because it allows us to study patterns of semantic change without relying on the continuity of specific lexemes, making it particularly useful for large-scale and long-term comparisons across languages and language families. It also enables us to treat meaning as a binary variable, since a pair of concepts can be colexified or not colexified at a given point in time. In addition, colexifications can be inferred from synchronic dictionary data, which is available for a large number of languages, and, unlike other outcomes of lexical change such as metaphor, do not require extensive knowledge of the individual languages and cultures.

\begin{figure}[htp]
    \centering
    \includegraphics[width=\textwidth]{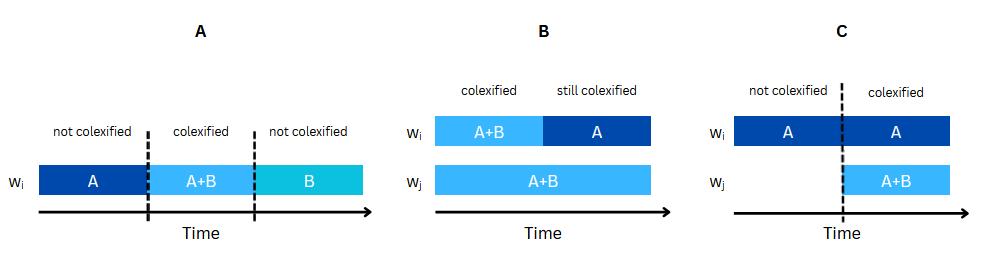}
    \caption{Interactions between colexification data and changes in meaning of words. Vertical dashed lines represent changes that are captured by our approach.}
    \label{fig:change}
\end{figure}

We follow a framework in which meaning change and colexification are linked: When a word shifts its meaning from A to B, there is an intermediate stage in which both meanings are expressed by the same form \parencite[133; Fig. \ref{fig:change}A]{campbell_historical_2020}. Thus, a change in colexification generally implies a lexical change. However, there are also cases where a change in the lexicon might go undetected if we only consider changes in the colexification of a concept pair. Imagine a situation in which meanings A and B are expressed by a word $w_i$ and thus colexified. Then, word $w_i$ loses meaning A, but there is another word $w_j$ that still expresses both meanings A and B. In this case, there is a meaning change in the lexicon, but no change in the colexification of A and B (Fig. \ref{fig:change}B). In another scenario, a word $w_i$ expressing the meaning A could be present in a language. Then, a word $w_j$ with meanings A and B is borrowed from another language. In this case, there is a change in colexification without a change in meaning in any of the existing words (Fig. \ref{fig:change}C). Nevertheless, the structure of the lexicon changes, and this change is captured by the colexification pattern.

Examples like this illustrate that there is no one-to-one correspondence between colexification patterns and meaning change. However, we assume that the default case, where a word shifts its meaning from A to B through an intermediate stage in which both meanings are colexified, is sufficiently common to allow inferences about lexical evolution from colexification patterns.

It is worth noting that not all colexifications arise from processes of lexical change. Pairs of unrelated concepts that share the same form due to sound change are known as \textit{homonyms} or \textit{homophones}. Such colexifications tend to be short lived, and specific homonyms typically occur only in a small number of languages, possibly due to a mechanism known as homophony avoidance \parencite{geeraerts1990homonymy, desmet2024who}. For example, while English \textipa{/s\textturnv n/} colexifies the concepts \textsc{sun} and \textsc{son}, no other Germanic language exhibits this colexification. To reduce the likelihood of including accidental homonyms, we therefore focus on concept pairs that are colexified across several languages within a family.

Previous studies on colexification have typically focused on factors that make concept pairs more likely to be colexified. Concepts that are closely related have been shown to be more likely to colexify across languages \parencite{xu_conceptual_2020, karjus_conceptual_2021}. However, \textcite{brochhagen_when_2022} show that the relationship between colexification and semantic relatedness follows a non-monotonic inverted U-shaped function, or Goldilocks curve, where an increase in relatedness raises colexification likelihood only up to a certain point. Beyond this point, relatedness becomes ``too high,'' and colexification likelihood decreases. For example, concept pairs that are very strongly semantically related (such as \textsc{Tuesday} and \textsc{Thursday}) show a lower propensity to colexify.

Another strand of research highlights that colexification patterns tend to follow certain constraints or conventions; for instance, it is cross-linguistically common to colexify a material and an artifact made from that material (e.g., the English word \textit{glass} denotes both a material and a drink vessel; \cite{srinivasan_how_2015}). Similarly, \textcite{youn_universal_2016} show that certain colexifications are very frequent across languages, independently of external factors such as geography or environment. Other studies have focused on colexification patterns in restricted semantic fields, such as emotion \parencite{jackson2019emotion}, body parts \parencite{tjuka2024universal}, and color terms \parencite{zaslavsky2018efficient}.

While the linguistic processes that give rise to these colexification patterns are inherently diachronic, few studies have explicitly examined the relationship between language change and colexification. \textcite{gast_patterns_2022} use data from European languages to show that colexification patterns are less persistent than the phonological forms of lexemes, indicating that meaning change occurs faster than lexical replacement. They also find that colexification patterns are more prone to borrowing than phonological matter, but only if one compares pairs of unrelated languages. Similarly, \textcite{urban_asymmetries_2011} investigates loose colexifications, concept pairs expressed by forms that are different but morphologically related, such as German \textit{sprechen} `speak' and \textit{besprechen} `discuss', to gain information about directionality in common colexifications, that is, which of the two meanings is original and which one is acquired. Finally, \textcite{carling_evolution_2023} use a phylogenetic method to reconstruct the meanings of proto-forms; in their approach, the different meanings that a proto-form evolves into create a diachronic colexification network.

In this study, we go beyond previous work and take full advantage of the opportunities offered by colexifications. We investigate the dynamics of lexical change using a large cross-linguistic dataset and apply phylogenetic comparative methods which model the underlying processes over time, account for genealogical relatedness, infer ancestral states, and detect biases and trends of change \parencite{greenhillLanguagePhylogeniesModelling2023a}. This approach offers a more comprehensive perspective on semantic evolution across extended time spans. To obtain robust and generalizable results, we investigate three language families from two different macroareas: Austronesian, Indo-European, and Uralic. This allows us to control, to some extent, for family- or region-specific outlier effects arising from historical, cultural, or environmental contingencies (see e.g., \cite{segerer2022areal}).

In section 2, we outline potential factors that may influence lexical change and how they are represented in our study. Sections 3 and 4 introduce the data and describe our modeling framework, including the phylogenetic comparative model used to infer change rates and long-term preferences. Section 5 presents the main results for the three language families, and section 6 discusses their implications for theories of semantic change.

\section{Dynamics of colexification}

We aim to identify the factors that drive long-term preferences in the organization of the lexicon, as evidenced by patterns of colexification. Using a phylogenetic model, we infer both stationary probabilities, i.e., the long-term probability that a concept pair is colexified within a language family, and speeds of change, which measure how often a concept pair changes from colexified to non-colexified, and vice versa. These parameters provide insight into how colexification patterns evolve over time. 
We build three lexical variables, (1) associativity, (2) frequency of use, and (3) borrowability, into our models of phylogenetic change in order to assess how different aspects of lexical change (stationary probability and speed) vary according to these predictors. 

%We link these properties of change to three lexical variables, , to assess how these predictors influence our measures of lexical change. 

%We then employ Bayesian regression models to assess how three lexical properties, associativity, frequency of use, and borrowability, influence our measures of lexical change.

\subsection{Associativity}
Associativity is a measure of the strength of the relationship between two words. We use the associativity between two words as a proxy for the semantic relatedness of the two concepts in a concept pair. Associativity rates are obtained from online experiments (e.g., \cite{de_deyne_better_2013}), in which participants are given a cue word and asked to produce an associated word. Associativity has been shown to predict the likelihood of colexification across languages \parencite{xu_conceptual_2020, brochhagen_when_2022}. Strongly associated concepts tend to be colexified in more languages. Therefore, we expect associativity to have a positive effect on the long-term preference for a colexification.

A similar measure of semantic relatedness shown to predict colexification likelihood is \textit{distributional similarity}, measured through word embeddings. However, both \textcite{xu_conceptual_2020} and \textcite{brochhagen_when_2022} have demonstrated that associativity is a stronger predictor of the number of languages that colexify a concept pair. For this reason, we do not investigate the role of distributional similarity in this study.

\subsection{Frequency of use}
Frequency of use has played an important role in studies of the lexicon at least since \textcite{zipf_human_1949}. Generally, word frequency is a measure of how familiar the average speaker is with a word \parencite{brysbaert2018word}. Frequency has been shown to affect lexical substitution: more frequent words are less often replaced than less frequent ones \parencite{pagel_frequency_2007, vejdemo_semantic_2016}. From this finding, we would generally expect a higher frequency of use to slow down lexical change. This is in line with \textcite{hamilton2016diachronic}, who show that more frequent words change their meanings, measured through embeddings, at a lower rate.

It has also been found that more frequent words tend to have more meanings \parencite{zipf_meaning-frequency_1945, casas_polysemy_2019}, which could indicate that higher frequency of use actually speeds up the acquisition of new meanings by a word. In that case, higher frequency might lead to a higher speed of change. On the other hand, words with more meanings could also simply be more frequent because they can be used in a wider range of situations.

Since we study changes in the colexification of specific concept pairs, and not changes of lexeme meanings, it is not entirely clear what effect frequency of use would have in our study. If higher frequency leads to the gain of several new meanings but not to the loss of previous ones, this would not necessarily be reflected in the rate of change of the colexification of a concept pair.

\textcite{xu_conceptual_2020} also tested whether usage frequency affects the likelihood that a concept pair is colexified across languages. Depending on the model they used, they found weakly positive or no effects. Therefore, we do not expect usage frequency to have a strong effect on the long-term preference for colexifying a concept pair.

\subsection{Borrowability}
Borrowability measures how prone a concept is to being expressed by a loanword. It has been shown that borrowability is negatively correlated with the speed of semantic change in cultural vocabulary \parencite{carling_evolution_2023} possibly because speakers perceive the connection between a borrowed lexeme and its meaning as stronger, treating it more like a proper noun than other lexemes. From this, we predict that higher borrowability of concepts corresponds to a lower change rate of colexifications. However, \textcite{carling_evolution_2023} also note that the pattern might differ for basic vocabulary as opposed to cultural vocabulary. Since our sample includes both types, the expected effect remains unclear.

\section{Data}
For our phylogenetic comparative model, we need three types of data. First, we need data pertaining to the response variable, i.e., concept pairs and their colexification status. Second, we need data on the predictor variables: associativity, frequency, and borrowability. Finally, we need language phylogenies that represent the relationships among the different languages within each language family. In the following, we describe how we collected and prepared the data for modeling.

\subsection{Concept pairs and colexifications}
We collect data on the colexification of concept pairs from the CLICS\textsuperscript{3} database \parencite{rzymski_database_2020}. CLICS\textsuperscript{3} is a dictionary database compiled from several cross-linguistic word lists, such as North\-Eura\-Lex \parencite{dellert_northeuralex_2020} and LexiRumah \parencite{edwards_lexirumah_2019}. The word forms in CLICS\textsuperscript{3} are linked to concepts from the Concepticon \parencite{List2016a}, a resource that harmonizes concept labels across different word lists. This allows forms from different languages or sources to be compared through their Concepticon glosses. Any concept set can be associated with more than one word form in a language, accounting for synonymy. CLICS\textsuperscript{3} contains data on 3,156 language varieties. However, some varieties appear in more than one word list and are thus counted multiple times (for instance, modern English is included in four word lists and therefore accounts for four out of the 3,156 varieties). We use data on 156 languages belonging to three language families: Austronesian, Indo-European, and Uralic. Languages are defined by their Glottocode in the CLICS\textsuperscript{3} data. Accordingly, two varieties with the same Glottocode are treated as the same language.

We consider a pair of concepts to be colexified in a language when at least one lexeme expresses both, even if there are more specific lexemes for each concept individually. For example, the German word \textit{Mann} expresses both the concepts \textsc{husband} and \textsc{man}, so we consider German to colexify these concepts. A more specific lexeme, \textit{Ehemann}, refers only to the concept \textsc{husband} but not \textsc{man}. However, because \textit{Mann} colexifies these concept, we consider this word irrelevant for our colexification judgment. We use different concept pairs for each of the three language families in our study, although there is considerable overlap. This approach accounts for the distinct cultural and linguistic characteristics of each family. 

To exclude words that are colexified due to homonymy, which reflect sound change rather than lexical change, we only consider concept pairs that are colexified in at least five languages within a given family.
We also exclude concepts that we consider grammatical (e.g., pronouns such as \textsc{we (inclusive)} or \textsc{they}) and concept pairs in which one or both of the concepts are attested in fewer than 30 languages in a family.

This filtering results in 273 concept pairs for Austronesian, 545 for Indo-European, and 84 for Uralic. The smaller number of Uralic pairs is likely due to a lower number of concepts per language and the smaller size of the Uralic family, which includes only 35 language varieties in the CLICS\textsuperscript{3} database, compared to 173 Indo-European and 395 Austronesian varieties.

\subsection{Predictors}

\subsubsection{Associativity scores}
Associativity refers to the semantic connection between two concepts often measured through word association tasks. Participants are presented with a cue word and instructed to provide a word they associate with it. For example, when given the cue \textit{arm}, a participant might respond with \textit{hand} or \textit{leg}. Our associativity measures come from the Small World of Words dataset for three resource languages: Dutch \parencite{de_deyne_better_2013}, English \parencite{de_deyne_small_2019}, and Rioplatense Spanish, a variety of Spanish spoken throughout most of Argentina and Uruguay \parencite{cabana_small_2024}. Because associativity is measured for words rather than concepts, we use words in specific languages as proxies for concept meanings.

The raw counts of responses are transformed into associativity scores using a random walk method, which has been shown to outperform other transformation methods \parencite{de_deyne_small_2019, brochhagen_when_2022}. To obtain associativity measures for the concept pairs in our study, we retrieve all forms for each concept in each concept pair from CLICS\textsuperscript{3} for the three resource languages and compute the associativity measures between each form of concept A and each form of concept B. We then take the arithmetic mean of these values to obtain the associativity score for each resource language, to then compute the arithmetic mean of these scores to derive a single associativity value for each concept pair.

We use the associativity values for English, Dutch, and Spanish words as a proxy for the associativity between two concepts. This approach does not account for the possibility of significant cultural variation in associativity scores. However, associativity measures from English and Dutch have been shown to predict colexification likelihood across a worldwide sample of languages \parencite{xu_conceptual_2020, brochhagen_when_2022}.

\subsubsection{Frequency scores}
Frequency of use is measured through corpus data. Specifically, we use the SUBTLEX corpus, a frequency database compiled from film subtitles. \textcite{new_use_2007} have shown that word frequencies obtained from subtitle corpora better predict participants’ performance in word recognition experiments than frequencies obtained from traditional book and newspaper corpora, possibly because they imitate naturalistic speech and therefore more accurately reflect how frequently words are used. However, these participants are often university students, and it is possible that other types of corpora might yield better results for different populations.

For simplicity, we collect data from the same resource languages used for the associativity measures, using SUBTLEX-US \parencite{brysbaert_moving_2009} for English, SUBTLEX-NL \parencite{keuleers_subtlex-nl_2010} for Dutch, and SUBTLEX-ESP \parencite{cuetos2012subtlex} for Spanish. We sum the frequencies of all word forms of a concept pair and then calculate so-called Zipf scores \parencite{van_heuven_subtlex-uk_2014} for each pair. Zipf scores are a logarithmic measure of the frequency at which a word appears in a corpus, generally ranging from 1 (0.01 occurrences per million words) to 8 (100,000 occurrences per million words).

Like associativity, frequency is measured for words rather than concepts; therefore, we use the aggregated frequencies of the words for a concept in CLICS\textsuperscript{3} for the resource languages as a proxy for the frequency of that concept.

\subsubsection{Borrowability scores}
Borrowability is the probability that a word for a specific concept is a loanword. We use data from the World Online Loanword Database (WOLD; \cite{haspelmath_loanword_2009}), which contains word lists for 1,460 meanings from a sample of 41 so-called recipient languages and records how many languages use a loanword to express a given meaning. Each word receives a ``borrowed score'' between 0 and 1, with 1 indicating ``clearly borrowed'' and 0 indicating ``no evidence for borrowing''. Intermediate scores, such as 0.75 for ``probably borrowed,'' are also used. An aggregate score for each meaning is then calculated based on the scores of all words expressing that meaning.

We mapped the WOLD meanings to Concepticon concepts to obtain a borrowability score for each concept in a concept pair. We then computed the mean of the two scores to obtain the borrowability score for the concept pair. If only one of the concepts had a borrowability score, we used that score.

We excluded concept pairs for which an associativity score, a frequency score, or a borrowability score was missing, keeping only concept pairs for which all three measures were available. This resulted in a final dataset of 140 concept pairs for Austronesian, 332 for Indo-European, and 84 for Uralic.

\subsection{Language phylogenies}
For the historical dimension of our study, we use phylogenetic trees for Austronesian \parencite{gray_language_2009}, Indo-European \parencite{heggarty_language_2023}, and Uralic \parencite{honkola_cultural_2013}. Phylogenetic trees model the historical relationships of languages by connecting languages that share a common ancestor and dating the split from that common ancestor. The trees we use were estimated using Bayesian phylogenetic methods, which produce a posterior sample of likely trees, i.e., likely histories of a language family. We matched the languages in the trees to the languages in the CLICS\textsuperscript{3} database by Glottocodes \parencite{hammarstrom_glottolog_2024}. The languages in each tree did not exactly match the languages for which we had colexification data, and our approach to handling this mismatch differed across the three families.

For Austronesian, many languages had substantial missing values (i.e., languages where at least one of the concepts of the concept pair is not attested in CLICS\textsuperscript{3}). This adds considerable uncertainty to the model. We therefore excluded all languages that had complete data for less than 100 of the 140 concept pairs. We also added languages to the tree for which we had complete data for at least 100 concept pairs. We did this using Glottolog trees \parencite{hammarstrom_glottolog_2024}, creating additional branches in our Bayesian trees that separated from the common ancestor of the most closely related language in the tree. If CLICS\textsuperscript{3} contained several varieties with the same Glottocode, we used the one with complete data for the most concept pairs, but replaced missing values with data from other varieties if available.

For Indo-European, we also added languages with a lot of complete data to the tree, but instead of using Glottolog trees, we added the languages manually to the most closely related branches, using uniform distributions to reflect uncertainty about the time of the split. 

For Uralic, there were relatively few missing values, since data on Uralic languages largely come from the same collections of word lists. We also added languages to the trees using information from Glottolog.

\section{Methods}

\subsection{Rate inference and regression}
\label{dist.phylo}

We use a standard phylogenetic comparative method for discrete data to characterize the diachronic dynamics of colexification.
We model changes in colexification over phylogenetic trees according to a continuous-time Markov process \parencite{Cathcart2018modeling}, under which frequencies of change between states of a character---in our case, concept pairs---are characterized by non-negative transition rates.
Since our characters have only two states, colexified (1) and non-colexified (0), transitions between states can be captured with a gain rate representing change from 0 to 1 and a loss rate representing change from 1 to 0. 

We use distributional phylogenetic modeling \parencite{cathcart_decoupling_2022,cathcart_rate_2023,cathcart2025linguistic}, which applies insights from distributional regression modeling \parencite{burkner2017bayesian} to phylogenetic methods, in order to assess the impact of different variables on patterns of change. 
This framework exploits the fact that gain and loss rates can be characterized according to a speed of change $s$ and a stationary probability $p$. For a given concept pair with index $i \in \{1,...,N\}$, the gain and loss rate are defined as follows: 
\begin{align*}
q^{\text{gain}}_i &= s_i \cdot p_i \\ 
q^{\text{loss}}_i &= s_i \cdot (1 - p_i)
\end{align*}
For each concept pair, it is possible to model $s_i$ and $p_i$ as functions of linear combinations of predictors, in the spirit of a regression framework, rather than directly sampling them from some prior distribution. These regression models estimate coefficients representing the effects of associativity, frequency, and borrowability on both the speed of change and the stationary probability. 
We use a logit link function to model stationary probability $p_i$:
\begin{equation}
    \text{logit}(p_i) = \beta^p_{0} + \beta^p_{\text{assoc}} \cdot \text{assoc}_i + \beta^p_{\text{freq}} \cdot \text{freq}_i + \beta^p_{\text{borrow}} \cdot \text{borrow}_i
\end{equation}
We use a log link function to model the speed of change $s_i$.
\begin{equation}
    \text{log}(s_i) = \beta^s_{0} + \beta^s_{\text{assoc}} \cdot \text{assoc}_i + \beta^s_{\text{freq}} \cdot \text{freq}_i + \beta^s_{\text{borrow}} \cdot \text{borrow}_i
\end{equation}
We fit four models, one in which both speed $s$ and stationary probability $p$ vary as functions of linear combinations of predictors (as described above), two restricted models, one where speed is modeled as a function of predictors while stationary probability is held constant, and vice versa, and a null model where both speed and stationary probability are held constant.
We carry out model comparison between the four models using Pareto-Smoothed Importance Sampling Leave-One-Out Cross-Validation (PSIS-LOO-CV; \cite{vehtari2017practical}).

For each family, parameter values were inferred jointly across concept pairs. We use normally distributed priors with a mean of 0 and a standard deviation of 1 for all regression coefficients and intercepts. For the restricted model that only models the relationships between the predictors and the stationary probability, we use a log-normal prior with a mean of 0 and a standard deviation of 1 over the speed of change. For the other restricted model, we use a uniform prior between 0 and 1 over the stationary probability. For the null model, we use both of these priors.
The likelihood of each individual concept pair is computed via Felsenstein's Pruning algorithm \parencite{felsenstein_maximum_1973}.

We infer posterior distributions of model parameters using the No-U-Turn Sampler (NUTS) in RStan \parencite{Carpenteretal2017}, using the {\tt map\_rect} function for parallelization. To incorporate phylogenetic uncertainty, we run models for 100 samples from the posterior tree sample for each family under study. We run models over 3 chains for 4000 iterations, discarding the first half of samples, resulting in 600,000 posterior samples per family. Convergence is monitored via the potential scale reduction factor \parencite{gelman1992inference}.

\subsection{Model Validation}
We use a recently introduced phylogenetic comparative method that integrates a continuous-time Markov process with regression models. 
Currently, this approach is not widely in use.
To ensure that results accurately reflect the data and are not artifacts of our model specification, we perform two validation procedures. First, we use simulated data to check whether the model can accurately recover known parameters. Second, we fit a negative binomial regression model for comparison.

\subsubsection{Simulation-based validation}
In the first validation procedure, we assess whether our distributional phylogenetic model can recover known parameters from simulated data. This allows us to determine the extent to which the results of the model might be artifacts of the model design. 
The data simulation process proceeds as follows: a coalescent phylogenetic tree comprising $L$ taxa (i.e., hypothetical languages) is randomly generated using the {\tt rcoal()} function from the package {\tt phytools} \parencite{phytools}. 
Then, synthetic predictor values $\boldsymbol x \in \mathbb{R}^D$ are randomly sampled from the standard normal distribution. 
The coefficients representing the effect of $\boldsymbol x$ on stationary probability and speed, $\beta^p_x,\beta^s_x$, are either fixed at $0$ or randomly sampled from the standard normal distribution, along with intercepts representing baseline values for stationary probability and speed. 
We then use the resulting gain and loss rates (generated as in \S\ref{dist.phylo}) to generate synthetic data on the phylogeny via a CTM process using the function {\tt sim.history()} from {\tt phytools}, yielding $D$ traits in $L$ taxa. 
Finally, we fit a distributional phylogenetic model to this synthetic dataset as described above, but using only one NUTS chain.

We carry out this procedure using 25 random number seeds, allowing the predictor variable $\boldsymbol x$ to have an effect on neither speed nor stationary probability, either speed or stationary probability, or both speed and stationary probability (yielding 4 different settings per random number seed), resulting in 100 simulations. 
We carry out this simulation procedure for synthetic data of different sizes, including a {\sc small} setting ($L=35,D=85$), a {\sc medium} setting ($L=75,D=85$), and a {\sc large} setting ($L=100,D=200$), roughly in line with the different amounts of data found across families used in our study. 
Following the simulation procedure, we consider a simulation to have accurately recovered a parameter value if the 95\% Equal-Tailed Interval (ETI) of inferred parameter values excludes zero (when the ground truth parameter is not zero) or includes zero (when the ground truth parameter is fixed at zero). 
We code inaccurately recovered parameters as false negatives (FN: if the ground truth parameter is nonzero but the 95\% ETI of inferred parameter values includes zero), false positives (FP: if the ground truth parameter is zero but the 95\% ETI of inferred parameter values excludes zero), and sign errors (SE: if the ground truth parameter is nonzero but the 95\% ETI of inferred parameter values includes only values on the other side of zero). 

\subsubsection{Negative binomial regression}
As the second validation step, we fit a negative binomial regression model using the \texttt{glm.nb()} function from the \texttt{MASS} package \parencite{venables2002MASS}. This allows us to examine the relationship between associativity, frequency, and borrowability on one side, and the number of languages that colexify a concept pair on the other. Negative binomial regression is appropriate for modeling count data, particularly when overdispersion is present, (i.e., when the variance exceeds the mean), making it a better fit than a standard Poisson model. In our case, the number of languages colexifying a given concept pair is assumed to follow a negative binomial distribution with mean $\mu_i$ and dispersion parameter $\theta$. The expected value $\mu_i$ is modeled using a log link function:
\begin{equation}
    \text{log}(\mu_i) = \beta_{0} + \beta_{\text{assoc}} \cdot \text{assoc}_i + \beta_{\text{freq}} \cdot \text{freq}_i + \beta_{\text{borrow}} \cdot \text{borrow}_i
\end{equation}
We use the same dataset and language sample as for the Bayesian models to allow for direct comparison.

\section{Results}

\subsection{Rate inference and regression}

The full model outperformed the restricted and null models in the Leave-One-Out Cross-Validation (Table \ref{elpdtable}). Compared to the full model, the expected log pointwise predictive density (ELPD) was lower by 26.67 (SE = 11.80) for the stationary-probability-only model, by 59.34 (SE = 21.25) for the speed-only model, and by 62.76 (SE = 20.34) for the null model. We therefore report only the coefficients estimated using the full model. Posterior distributions of the parameters are summarized by their medians and 95\% equal-tailed intervals (ETIs), defined as the range between the 2.5\% and 97.5\% quantiles.

\begin{table}[htp]
\centering
%\begin{tabular}{lc}
% Model & \multicolumn{1}{l}{ELPD difference} \\[0.2ex]
% \hline\\[-1.8ex]
%Full model & 0 \\[2ex]
%\begin{tabular}[c]{@{}l@{}}Stationary-\\ probability-only\end{tabular} & \begin{tabular}[c]{@{}c@{}}-26.67\\ (11.80)\end{tabular} \\[2ex]
%Speed-only & \begin{tabular}[c]{@{}c@{}}-59.34\\ (21.25)\end{tabular} \\[2ex]
%Null model & \begin{tabular}[c]{@{}c@{}}-62.76\\ (20.34)\end{tabular}
%\end{tabular}

\begin{tabular}{ll}
Model & ELPD difference\\
\hline
Full model & 0 \\
Stationary probability-only & -26.67 (11.80)\\
Speed-only & -59.34 (21.25)\\
Null model & -62.76 (20.34)\\
\end{tabular}

\caption{Differences in expected log pointwise predictive densities (ELPD) between the four models, relative to the model with the highest ELPD, the full model. Standard errors in parentheses.}
\label{elpdtable}
\end{table}

\begin{figure}[htp]
    \centering
    \includegraphics[width=\textwidth]{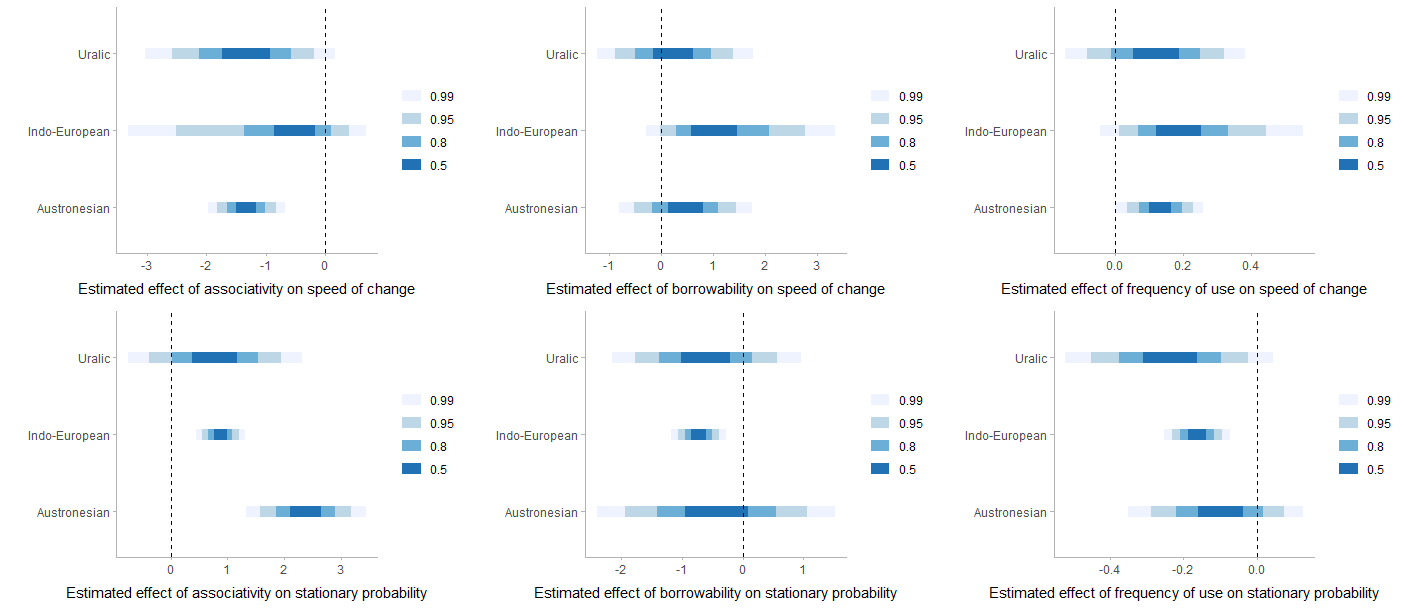}
    \caption{Posterior samples of the regression coefficients}
    \label{fig:plot}
\end{figure}

Estimated coefficients are presented in Table \ref{coeftable} and visualized in Fig. \ref{fig:plot}. Associativity showed a positive effect on the stationary probability of a colexification. The 95\% ETI lay entirely above zero for Austronesian (median = 2.38, 95\% ETI: [1.57, 3.19]) and Indo-European (median = 0.87, 95\% ETI: [0.55, 1.20]), but not for Uralic (median = 0.78, 95\% ETI: [-0.38, 1.94]). Conversely, associativity had a negative effect on the speed of change of a colexification, indicating that more closely associated concept pairs tend to form more stable colexifications. Here, the 95\% ETI was entirely below zero for Austronesian (median = -1.33, 95\% ETI: [-1.82, -0.84]) and Uralic (median = -1.33, 95\% ETI: [-2.58, -0.19]), but not for Indo-European (median = -0.46, 95\% ETI: [-2.69, 0.39]).

\begin{table}[htp]
\centering
\begin{tabular}{llccc}
 Parameter & Language family & \multicolumn{1}{l}{Associativity} & \multicolumn{1}{l}{Borrowability} & \multicolumn{1}{l}{Frequency} \\[0.5ex] \hline
 & Uralic & \begin{tabular}[c]{@{}c@{}}-1.33\\ {[}-2.58, -0.19{]}\end{tabular} & \begin{tabular}[c]{@{}c@{}}0.23\\ {[}-0.88, 1.38{]}\end{tabular} & \begin{tabular}[c]{@{}c@{}}0.12\\ {[}-0.08, 0.32{]}\end{tabular} \\ [2ex]
Speed of change & Indo-European & \begin{tabular}[c]{@{}c@{}}-0.46\\ {[}-2.69, 0.39{]}\end{tabular} & \begin{tabular}[c]{@{}c@{}}0.91\\ {[}-0.02, 2.72{]}\end{tabular} & \begin{tabular}[c]{@{}c@{}}0.17\\ {[}0.003, 0.45{]}\end{tabular} \\ [2ex]
 & Austronesian & \begin{tabular}[c]{@{}c@{}}-1.33\\ {[}-1.82, -0.84{]}\end{tabular} & \begin{tabular}[c]{@{}c@{}}0.47\\ {[}-0.51, 1.44{]}\end{tabular} & \begin{tabular}[c]{@{}c@{}}0.13\\ {[}0.04, 0.23{]}\end{tabular} \\ [2ex] \hline
 & Uralic & \begin{tabular}[c]{@{}c@{}}0.78\\ {[}-0.38, 1.94{]}\end{tabular} & \begin{tabular}[c]{@{}c@{}}-0.61\\ {[}-1.77, 0.57{]}\end{tabular} & \begin{tabular}[c]{@{}c@{}}-0.24\\ {[}-0.45, -0.02{]}\end{tabular} \\ [2ex]
Stationary probability & Indo-European & \begin{tabular}[c]{@{}c@{}}0.87\\ {[}0.55, 1.20{]}\end{tabular} & \begin{tabular}[c]{@{}c@{}}-0.72\\ {[}-1.07, -0.38{]}\end{tabular} & \begin{tabular}[c]{@{}c@{}}-0.16\\ {[}-0.23, -0.09{]}\end{tabular} \\ [2ex]
 & Austronesian & \begin{tabular}[c]{@{}c@{}}2.38\\ {[}1.57, 3.19{]}\end{tabular} & \begin{tabular}[c]{@{}c@{}}-0.43\\ {[}-1.92, 1.05{]}\end{tabular} & \begin{tabular}[c]{@{}c@{}}-0.10\\ {[}-0.29, 0.07{]}\end{tabular}
\end{tabular}
\caption{Median and 95\% Equal-Tailed Interval of the posterior distributions of regression coefficients.}
\label{coeftable}
\end{table}

Frequency of use exhibited a negative effect on stationary probability, suggesting that colexifications involving more frequently used concepts are less preferred. This effect was supported for Indo-European (median = -0.16, 95\% ETI: [-0.23, -0.09]) and Uralic (median = -0.24, 95\% ETI: [-0.45, -0.02]), but not for Austronesian (median = -0.10, 95\% ETI: [-0.29, 0.07]). Frequency also had a positive effect on the speed of change, indicating that colexifications of more frequently used concepts tend to be less stable. Here, 95\% ETIs were entirely above zero for Austronesian (median = 0.13, 95\% ETI: [0.04, 0.23]) and Indo-European (median = 0.17, 95\% ETI: [0.003, 0.45]), but not for Uralic (median = 0.12, 95\% ETI: [-0.08, 0.32]).

Finally, borrowability had a clear negative effect on stationary probability only for Indo-European (median = -0.72, 95\% ETI: [-1.07, -0.38]), but not for Austronesian (median = -0.43, 95\% ETI: [-1.92, 1.05]) or Uralic (median = -0.61, 95\% ETI: [-1.77, 0.57]). No clear effects of borrowability were observed on the speed of change, as the 95\% ETI included zero for all three language families: Indo-European (median = 0.91, 95\% ETI: [-0.02, 2.72]), Austronesian (median = 0.47, 95\% ETI: [-0.51, 1.44]) or Uralic (median = 0.23, 95\% ETI: [-0.88, 1.38]).

\subsection{Model validation}

\subsubsection{Simulation-based validation}
Results of the simulation procedure are presented in Table \ref{tab:simulations}. As shown, ground truth parameter values are not recovered 100\% of the time, with accuracy ranging from 80\% to 94\%. Most errors are false negatives, with under 5\% of simulated parameter values yielding false positives across all simulation settings. Sign errors are nonexistent. This indicates that our model fitting procedure is somewhat conservative, but not overly so.

\begin{table}[htp]
    \caption{Summary of simulation-based validation procedure for different synthetic data sizes for parameters $\beta^s_x$ and $\beta^p_x$. Parameters are recovered accurately (T) or as false positives (FP), false negatives (FN), and sign errors (SE).}
    \label{tab:simulations}
    \centering
    \begin{tabular}{l|c|c|c|c}
        Parameter & T & FP & FN & SE\\
        \hline
        $\beta^s_x$, {\sc small} & 85 & 2 & 13 & 0\\
        $\beta^p_x$, {\sc small} & 86 & 4 & 10 & 0\\
        \hline
        $\beta^s_x$, {\sc medium} & 80 & 3 & 17 & 0\\
        $\beta^p_x$, {\sc medium} & 91 & 1 & 8 & 0\\
        \hline
        $\beta^s_x$, {\sc large} & 87 & 4 & 9 & 0\\
        $\beta^p_x$, {\sc large} & 94 & 0 & 6 & 0\\
        \hline
    \end{tabular}
\end{table}

\subsubsection{Negative binomial regression}
Results of the negative binomial regression model are presented in Table \ref{negbin}. Associativity shows a positive effect on the number of colexifying languages in Austronesian and Indo-European, but not in Uralic. Frequency has a significantly negative effect only in Uralic, whereas borrowability shows a negative effect only in Indo-European.

\begin{table}[!htbp] \centering 
  \caption{Results of the negative binomial regression model} 
  \label{negbin} 
\begin{tabular}{@{\extracolsep{5pt}}lccc} 
\\[-1.8ex]\hline 
\hline \\[-1.8ex] 
 & \multicolumn{3}{c}{\textit{Dependent variable:}} \\ 
\cline{2-4} 
\\[-1.8ex] & \multicolumn{3}{c}{Number of colexifying languages} \\ 
\\[-1.8ex] & Austronesian & Indo-European & Uralic\\ 
\hline \\[-1.8ex] 
 Associativity & 1.704$^{***}$ & 0.679$^{***}$ & 0.366 \\ 
  & (0.517) & (0.218) & (0.515) \\ 
  & & & \\ 
 Frequency & 0.022 & $-$0.051 & $-$0.179$^{*}$ \\ 
  & (0.114) & (0.049) & (0.094) \\ 
  & & & \\ 
 Borrowability & $-$0.576 & $-$0.717$^{***}$ & $-$0.643 \\ 
  & (1.207) & (0.239) & (0.508) \\ 
  & & & \\ 
 Constant & 1.894$^{***}$ & 2.536$^{***}$ & 3.368$^{***}$ \\ 
  & (0.712) & (0.286) & (0.569) \\ 
  & & & \\ 
\hline \\[-1.8ex] 
Observations & 140 & 332 & 84 \\ 
Log Likelihood & $-$471.230 & $-$1,015.503 & $-$262.057 \\ 
$\theta$ & 1.095$^{***}$  (0.141) & 4.712$^{***}$  (0.508) & 4.832$^{***}$  (1.030) \\ 
Akaike Inf. Crit. & 950.460 & 2,039.005 & 532.114 \\ 
\hline 
\hline \\[-1.8ex] 
\textit{Note:}  & \multicolumn{3}{r}{$^{*}$p$<$0.1; $^{**}$p$<$0.05; $^{***}$p$<$0.01} \\ 
\end{tabular} 
\end{table}

\clearpage

\section{Discussion}
Using a phylogenetic model, we tested the effect of three predictors on the change rates and stationary probabilities of commonly colexified concept pairs. We observed several potentially meaningful effects; however, none reached a 95\% probability threshold across all three language families.

The positive effect of associativity on the stationary probability meaning the long-term preference for a colexification, extends previous work \parencite{xu_conceptual_2020, brochhagen_when_2022} by showing how these preferences arise over time through evolutionary processes. It is somewhat surprising, however, that this relationship is not clearly observed in the Uralic language family. The lack of a clear pattern may result from the smaller number of colexifications and languages in the Uralic dataset, which lead to higher variance in the posterior distribution, as seen in Fig. \ref{fig:plot}. The negative binomial model yields similar results, with associativity having a positive effect on the number of languages that colexify a given concept pair. 
It should also be noted that we only included attested colexifications in our data, whereas previous studies such as \textcite{xu_conceptual_2020} and \textcite{brochhagen_when_2022} also sampled concept pairs that were never colexified. These non-colexified pairs are likely to have very low associativity scores, as two randomly selected words have only a small chance of being associated with each other. Including these low-associativity, low-colexification concept pairs might lead to a stronger, more pronounced effect, but it is difficult to reconcile with the phylogenetic models we use.

Similarly, the negative effect of associativity on change rates confirms previous findings. Concept pairs with high associativity tend to form more stable colexifications, meaning they are rarely lost once they appear. Because these concepts are so closely related, speakers who colexify them might not see them as distinct. For example, the concepts \textsc{knife} and \textsc{knife (for eating)} are colexified in 282 out of 328 language varieties that have a CLICS\textsuperscript{3} entry for \textsc{knife (for eating)}, and speakers may treat these as the same thing. By contrast, colexified concepts that are less strongly associated might still be understood as different. For example, German colexifies \textsc{sharp} and \textsc{spicy} through the word \textit{scharf}, but it seems less likely that German speakers perceive these as being the same meaning. Such colexifications may be more likely to disappear if one meaning is replaced, while pairs like \textsc{knife} and \textsc{knife (for eating)} tend to persist. This explanation is, however, speculative, and the mechanisms behind it need further study.

Frequency of use has a positive effect on the speed of change, indicating that colexifications of more frequently used concepts are gained or lost faster than those of less frequently used concepts. This might be related to findings that words for commonly used concepts are less often substituted \parencite{pagel_frequency_2007, vejdemo_semantic_2016}, so their forms persist for a longer time and have more opportunity to gain or lose meanings, increasing the chances for colexifications. However, this would only imply that there are many changes in the meanings of a particular word form, not in a particular colexification. It still remains unclear why the change rate of specific colexifications would be affected by a word form having many meanings. It is possible that a form acquiring too many meanings is undesirable, and that speakers counter this by substituting forms for some of the meanings, resulting in a loss of colexifications.

%\todo[inline]{I think this result is going to be quite challenging to reconcile with the general finding in the literature that more frequent words tend to be more polysemous and gain more senses (e.g., issues raised in Hamilton et al. 2016 and lit cited). 

%Some reviewers may be concerned that the effect we are getting is an artifact of our model setup. Can we show that within our data, frequency is synchronically negatively associated with colexification?
%}

Related to this, colexifications of more frequently used concepts seem to be less preferred over time, contrary to the findings of \textcite{xu_conceptual_2020}. A possible explanation is that, because frequently used concepts are more salient in speakers' minds, they are less constrained to colexify exclusively with concepts that they are strongly associated with. This may result in frequently used concepts participating in many colexifications that are less common and frequently gained and lost rather than forming a few cross-linguistically common colexifications that persist over time.

We obtain somewhat similar results from the negative binomial model, although the negative effect of frequency is only significant for the Uralic languages. However, this model considers only raw colexification counts and does not account for the genetic relationships between languages. As a result, if several closely related languages colexify a concept pair, their influence on the negative binomial model will be stronger than in the Bayesian model.

It should be noted that the frequency scores obtained from subtitle corpora of English, Dutch, and Spanish may be unreliable. In particular, it is unclear to what extent the frequencies in corpora of scripted ``staged orality'' \parencite{heyd2010, berbersardinhaVeiranopinto2019} are representative of general familiarity. Such frequencies may differ substantially from frequencies of actual use and perception, which arguably reflect and shape individual and population-wide familiarity with concepts more accurately. While these corpora are a good measure of how familiar contemporary university students are with certain concepts \parencite{brysbaert_moving_2009}, they might not accurately reflect other populations. In particular, the familiarity of past societies with cultural terms might have been considerably different from that of Western European societies today. For example, a society in which most people were farmers might have used concepts such as \textsc{sow (female pig)} much more frequently than contemporary English speakers. The same applies to modern cultures that differ fundamentally from Western Europe. However, many of the concepts in our study are not part of cultural vocabulary, and speakers’ familiarity with them may not have varied greatly over time or across languages.

In contrast to frequency, a clear effect of borrowability on the development of colexifications was only found for the Indo-European language family. The results show a negative correlation between borrowability and the long-term preference for a colexification. This negative relationship is also observed in the negative binomial model. A possible explanation is that borrowing can disrupt existing colexifications when a lexeme is borrowed that expresses only one of two previously colexified concepts, leading to the loss of that colexification. Conversely, borrowing could also result in the formation of a new colexification if a lexeme expressing two meanings is adopted from another language. However, our results suggest that this is comparatively rare. It should also be noted that borrowing scores from the WOLD may overestimate borrowing rates, as many borrowings into non-Indo-European languages stem from contact with dominant Indo-European languages during the colonial period \parencite{carling2019causality}. Languages that were less affected by colonization therefore tend to show lower-than-average borrowing rates.

Furthermore, we find considerable differences between the three language families. Although we never estimate opposing effects (e.g., a positive effect in one family and a negative effect in another), none of the relationships we test can be demonstrated consistently across all families. This may be partly due to the fact that we included a different set of concept pairs for each language family. Which colexifications are common within a family could depend on cultural factors, but also on the CLICS\textsuperscript{3} data itself, which may contain distinct sets of concepts for different languages. Another possibility is that the effects we observe are not uniform across colexifications. In this case, one family might include common colexifications for which a given effect is strong, while in another family the more frequent colexifications might show weaker effects.

We also observe clear differences in the width of the credible intervals across language families. For the effects on the stationary probability, Indo-European shows narrow intervals, indicating low uncertainty in the estimated coefficients, while the other two families display much broader distributions. This suggests that the effects are more consistently expressed within Indo-European languages, whereas the patterns in Uralic and Austronesian are more variable. One reason for this might be the larger number of concept pairs and languages in the Indo-European dataset, which provides stronger statistical support. Another possibility is that Indo-European colexifications exhibit a clearer signal, with fewer cases that deviate from general trends; for instance, few concept pairs that combine low associativity with high stationary probability. This could indicate that the underlying mechanisms shaping colexifications are more homogeneous within Indo-European, while they vary more across the other families.

On the other hand, for the effects on the speed of change, Indo-European shows wider credible intervals than the other two families. A high speed of change indicates that closely related languages often differ in whether they exhibit a particular colexification. In other words, a colexification is characterized by a high speed of change when many pairs of closely related languages differ in its presence or absence. The greater uncertainty in the Indo-European estimates could therefore reflect higher variability in the data; that is, some colexifications remain stable across related languages, showing a low speed of change, while others vary substantially and change more frequently.

While our results reveal meaningful patterns, they should also be interpreted in light of certain limitations of our data and modeling framework. One such limitation is that we did not model language contact. This is relevant because some colexification patterns have been shown to be prevalent in specific areas \parencite{schapper_baring_2022, segerer2022areal} and tend to be borrowed at a higher rate compared to lexical form \parencite{gast_patterns_2022}. In addition, our study is restricted to concept pairs that are commonly colexified across languages. While this allows us to model meaning change for some concept pairs (for example, when a word for \textsc{man} gains the meaning \textsc{husband}, and vice versa), it may not be possible to generalize our results to meaning changes between rarely colexified concepts (such as a word for \textsc{man} gaining the meaning \textsc{boy}). Finally, as discussed above, our data for frequency and associativity are based on Indo-European languages originally spoken in Europe and may therefore not fully capture cross-linguistic variation.

Overall, our results show both general tendencies and differences between language families. Some predictors, like associativity and frequency, have clear effects in certain families, but these effects are not consistent across all families. This suggests that lexical change is influenced by a combination of general tendencies and family- or culture-specific factors. At the same time, variation in the data and differences in colexification coverage highlight the need for further studies with larger and more diverse samples.

%In sum, we have demonstrated that our understanding of semantic change and the forces that drive it can be substantially advanced through the use of synchronic colexification data combined with temporally explicit models of language change. Our approach broadens the empirical and temporal scope of previous work, offering wider coverage of languages, language families, and data, independence from morphological matter, cross-linguistic applicability, and controls for phylogenetic correlation, and, to some extent, cultural differences. Our results provide only partial support for previous findings and claims. In particular, our findings challenge the assumption that results obtained from small language samples and rather short time spans are representative of the long histories of culturally and linguistically diverse language families.

\section{Conclusions}

%In this paper, we have demonstrated that our understanding of semantic change and the forces that drive it can be substantially advanced through the use of synchronic colexification data combined with temporally explicit models of language change. 

In this paper, we have obtained  results that provide only partial support for previous findings and claims. In particular, the effects of associativity, frequency, and borrowability differ across the three language families under study, suggesting that the organization of the lexicon is shaped by both general tendencies and family-specific factors. These results challenge the assumption that findings based on small samples or limited time spans can be generalized to the histories of culturally and linguistically diverse language families. They also illustrate that large-scale phylogenetic modeling can uncover long-term biases in semantic evolution that are not always visible from synchronic data alone.

At the same time, our findings must be interpreted in light of several limitations. The data rely on cross-linguistically common colexification patterns, which may not capture all semantic nuances or rare colexifications. Moreover, our analyses focus on only three language families, and the values for frequency and associativity are solely based on Indo-European sources. These factors point to the need for more comprehensive datasets that go beyond the languages and cultural contexts most readily available.

Future work can build on this framework in several ways. A key next step is to explicitly model contact and borrowing processes, since the structure of the lexicon often diffuses areally and transcends genealogical boundaries. Expanding the range of semantic domains will further improve our understanding of the mechanisms shaping meaning change.

%Taken together, this study demonstrates that colexification data, when analyzed within an evolutionary modeling framework, offer a powerful tool for tracing the dynamics of lexical meaning at the level of established language families. 

Taken together, we have demonstrated that our understanding of semantic change and the forces that drive it can be substantially advanced through the use of synchronic colexification data combined with temporally explicit models of language change. Our approach broadens the empirical and temporal scope of previous work, offering wider coverage of languages, language families, and data, independence from morphological matter, cross-linguistic applicability, and controls for phylogenetic correlation and, to some extent, cultural differences. This novel approach opens new possibilities for uncovering principles behind lexical change.

%\section*{Acknowledgments}

\section*{Supplementary Materials}
The data and code we used for this study can be accessed at \url{https://doi.org/10.5281/zenodo.17302972}.

\section*{Acknowledgments}
We are grateful to Gerd Carling and Laura Dees for helpful feedback, as well as attendees of the Phylogenetic Modeling and Human History Workshop held June 6--7, 2024 at the University of Zurich. 
C.C. thanks Dan Dediu for providing Stan code implementing numerically stable matrix exponentiation. 
K.G. was supported by a Gerda Zeltner fellowship. 
C.C. was supported by the NCCR Evolving Language (SNSF Agreement No.\ 51NF40\underline{\phantom{x}}180888). %\todo{not sure if this funding acknowledgment is correct: feel free to edit}

\section*{Author Contributions}
\begin{description}
    \item K.G.: Conceptualization, Methodology, Software, Validation, Formal analysis, Data Curation, Writing -- Original Draft, Writing -- Review \& Editing, Visualization
    \item S.S.: Conceptualization, Writing -- Review \& Editing, Supervision, Funding acquisition
    \item C.C.: Methodology, Software, Validation, Formal analysis, Writing -- Original Draft, Writing -- Review \& Editing
    \item P.W.: Conceptualization, Methodology, Writing -- Review \& Editing, Supervision, Funding acquisition
\end{description}

\printbibliography

\end{document}